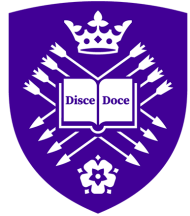

# University of Sheffield

# Yantra AI - An intelligence platform which interacts with manufacturing operations

Varshini Krishnamurthy

Supervisor: Andrew Stratton

*A report submitted in fulfilment of the requirements*
*for the degree of MSc in Data Analytics*

*in the*

*Department of Computer Science*

September 11, 2024

# Declaration

All sentences or passages quoted in this report from other people's work have been specifically acknowledged by clear cross-referencing to author, work and page(s). Any illustrations that are not the work of the author of this report have been used with the explicit permission of the originator and are specifically acknowledged. I understand that failure to do this amounts to plagiarism and will be considered grounds for failure in this project and the degree examination as a whole.

Name: Varshini Krishnamurthy

Signature: Varshini Krishnamurthy

Date: 11th September 2024

# Abstract

Industry 4.0 is growing quickly, which has changed smart production by encouraging the use of real-time tracking, machine learning, and AI-driven systems to make operations run more smoothly. The main focus of this dissertation is on creating and testing an intelligent production system for XRIT that solves important problems like energy management, predictive maintenance, and AI-powered decision support. Machine learning models are built into the system, such as the Random Forest Classifier for proactive maintenance and the Isolation Forest for finding outliers. These models help with decision-making and reducing downtime. Streamlit makes real-time data visualisation possible, giving workers access to dashboards that they can interact with and see real-time observations.The system was tested with fake data and is made to be scalable, so it can be used in real time in XRIT's production setting. Adding an AI-powered virtual assistant made with GPT-4 lets workers get real-time, useful information that makes complicated questions easier to answer and improves operational decisions. The testing shows that the system makes working efficiency, energy management, and the ability to plan repairs much better. Moving the system to real-time data merging and looking for other ways to make it better will be the main focus of future work.

# Acknowledgements

I want to thank my supervisor, Andrew Stratton, from the bottom of my heart for all the help, advice, and direction he gave me during this project. His knowledge and support have been very important to the development of my studies and career.

I'd also like to thank Viswa and Andrew from XRIT for giving me the chance to work on smart production systems, predictive maintenance, and AI-driven insights.

I'm very thankful to my family and friends for always supporting me and to my coworkers and peers for the helpful comments that made this study much better.

Thanks to everyone who helped.

# Acronym

AI - Artificial Intelligence
PdM - Predictive Maintenance
SVM - Support Vector Machines
XAI - Explainable Artificial Intelligence
XSS - Cross-Site Scripting
SQL - Structured Query Language
IoT - Internet of Things
IF - Isolation Forest
NLP - Natural Language Processing
MSc - Master of Science
KPIs - Key Performance Indicators
XRIT - The name of the company (not an acronym, but a project-related entity)
DES - Discrete Event Simulations
AGV - Automated Guided Vehicle

# Contents

# List of Figures

# Chapter 1: Introduction

Because of the use of advanced technologies like artificial intelligence (AI), real-time data, and machine learning, many businesses have changed a lot since Industry 4.0 came out. These tools must be used by businesses that want to be more competitive, run their processes more efficiently, and make better use of their resources. One of the hardest things for many businesses today is avoiding unplanned breaks, which can seriously slow down production and cost a lot of money. Energy management that works well has also become more important as businesses try to find a balance between being productive and being environmentally friendly.

Due to these problems in the industry, XRIT, a business with many clients, is looking into using smart solutions to make its operations run more smoothly. The main goal of this project is to create an AI-powered system that uses machine learning models to control energy, do predictive maintenance, and find problems in real time. A Random Forest Classifier predicts when equipment might break down, an Isolation Forest finds strange things in operating data streams, and a RandomForest Regressor predicts how much energy will be used. Together, these models make it easier for proactive actions, reduce downtime, make the best use of resources, and improve total operating performance.

The system is designed with scalability and adaptability in mind, beginning with model validation through simulated data before transitioning to real-time deployment. This step-by-step process makes sure that the system can handle the complexity of real-life business settings while still being reliable and effective. The project supports Industry 4.0's main goal of using new technologies to improve business processes by encouraging smart automation and letting people make decisions based on data.

This project uses a lot of the things I learned and learned how to do while I was getting my MSc in Data Analytics. Specifically, it uses machine learning, real-time data handling, and AI-based decision-making tools. Using these methods in a real-life industry setting has shown how hard it is to make systems bigger while still making sure the models are correct. It was easy to understand the theory ideas, but this project has shown how hard it is to keep a system working the same way when practical conditions change quickly.

The main goals of this project are to create a system for predictive maintenance that can predict when equipment will break down, cut down on unplanned downtime, and allow for quick maintenance interventions.

- Putting in place a real-time anomaly detection system to find problems in practical data streams and make fixes quickly.
- Using machine learning to correctly predict how much energy will be used will lead to better resource management and less energy being used than is necessary.
- Adding an AI-powered virtual helper will give workers real-time information that will make making decisions easier.

Several problems came up during the project. One of the main problems was that models were built using artificial data, which doesn't fully represent how unpredictable real-world settings are. When we switch to real-time data, we will have to keep fine-tuning the model to make sure the system is stable. A big task was also making sure the system could grow as XRIT did. The system must be able to handle more and more data while still being accurate and using computers efficiently. Finally, making a user interface that was easy to understand and use was important to make sure that the AI-powered virtual helper could give workers useful information.

# Chapter 2: Literature Review

## 2.1 Introduction

The following section takes a close look at the literature that is related to the creation of the smart production system that is being used in this project. It works on important areas of integrating technology, like watching data in real time, planning maintenance ahead of time, finding oddities, improving processes, and AI-powered virtual assistants. In each section, we look at the latest study in these areas, comparing different approaches and figuring out how this project adds to or changes previous work. The chapter also points out gaps in the current research and talks about how the suggested method plans to fill these gaps.

## 2.2 Real-Time Monitoring and Predictive Maintenance in Smart Manufacturing References

In modern industrial systems, real-time monitoring is critical for maintaining optimal performance and minimising downtime. Predictive maintenance, driven by big data pipelines, emphasises the integration of data from diverse sources such as sensors and controllers to support data-driven maintenance strategies [1]. These systems allow manufacturers to process real-time data, improving efficiency and reducing machine downtime [1]. By employing predictive analytics, maintenance activities can be scheduled based on real-time and historical data, ensuring system uptime and minimising manufacturing disruptions [1].

In the context of Industry 4.0, real-time production scheduling plays a crucial role in managing uncertainties, such as unexpected machine breakdowns or job arrivals. Dynamic rescheduling models, which update in real time, are essential for adapting to these disruptions, thereby enhancing machine availability and overall production efficiency [2]. The system developed in this project aligns with these principles by utilising predictive analytics to enable proactive machine maintenance and continuous system monitoring, thereby reducing the risk of operational failures [2].

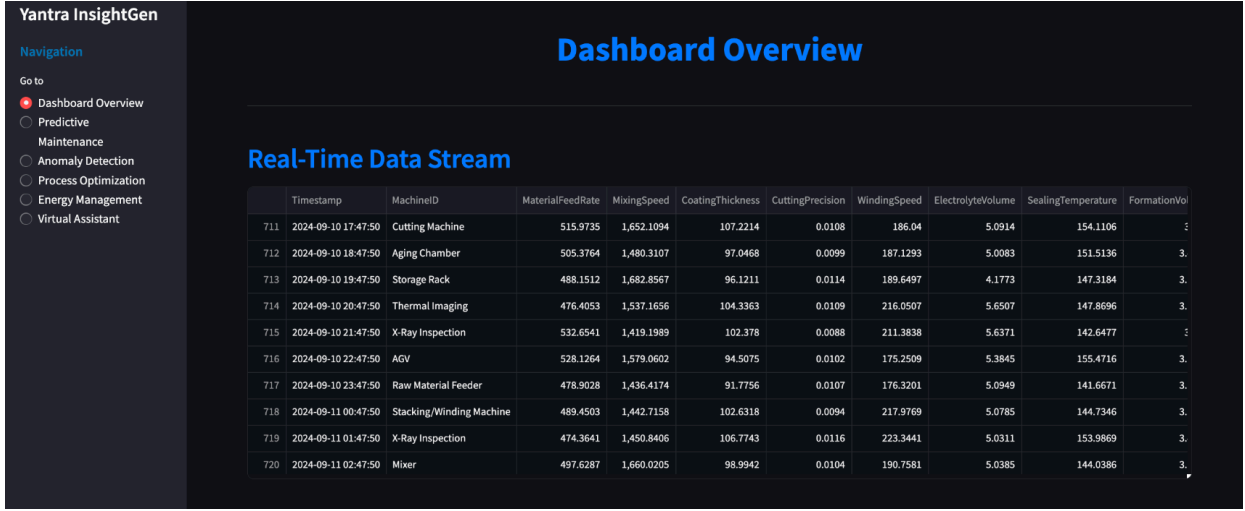

*Figure 2.1: Dashboard Overview for Real-Time Data Stream Monitoring*

Technologies for Data Monitoring

Technologies that facilitate real-time data monitoring are pivotal in Industry 4.0, providing live visibility into operational data streams and enabling timely decision-making. Real-time systems typically consist of both hardware, such as sensors and actuators, and software frameworks that process and visualise data. Tools like Streamlit have gained prominence as frameworks for real-time data visualisation in industrial settings, offering customizable, interactive dashboards [3]. These dashboards allow operators to track key performance metrics in real time, facilitating immediate responses to anomalies or inefficiencies [3]. The ability of Streamlit to seamlessly integrate with data sources and machine learning models makes it particularly effective for smart manufacturing applications, where real-time feedback on production metrics is essential [3].

By coupling real-time monitoring with predictive maintenance systems, manufacturers can further reduce downtime and increase system reliability [4]. This integration enables operators to identify potential issues before they escalate into major problems, enhancing the overall efficiency of the production process. The system developed in this project incorporates these technologies, using real-time data to support predictive maintenance and ensure uninterrupted operations [4].

Challenges in Data Monitoring

While real-time data monitoring is a critical component of modern manufacturing systems, several challenges can limit its effectiveness. A significant challenge is the integration of heterogeneous data sources, which may involve different data formats, communication protocols, and varying latencies. This complexity requires robust systems capable of harmonising and synchronising data streams to ensure accurate and timely insights [5]. Additionally, data reliability is a concern, as sensor noise, outliers, and missing data can compromise the quality of the insights provided. These issues may lead to incorrect conclusions or inefficiencies within the system [5].

Scalability is another persistent challenge in large-scale industrial environments, where vast amounts of data are generated. Systems must process high data throughput without performance degradation, as delays in data processing can negate the advantages of real-time monitoring [5]. Furthermore, cybersecurity risks, particularly in systems involving interconnected IoT devices and cloud platforms, present a significant threat to data integrity and operational continuity [5]. The system proposed in this project addresses these challenges by employing robust data processing techniques, ensuring reliable real-time monitoring while mitigating risks associated with data integration and cybersecurity.

## 2.3 Predictive Maintenance Using Machine Learning

Overview of Predictive Maintenance

Predictive maintenance (PdM) is a data-driven approach that utilises advanced monitoring techniques to predict equipment failures before they occur, thereby minimising unplanned downtime and improving machine reliability. PdM continuously gathers data from machinery, analysing it to detect patterns that indicate potential issues, enabling timely intervention. This proactive approach allows for maintenance activities to be scheduled based on the actual condition of equipment rather than adhering to fixed schedules or responding reactively to breakdowns [6]. In industrial settings, predictive maintenance has been shown to extend the operational life of machines, reduce maintenance costs, and improve overall production efficiency [6].

Random Forest Classifier for Predictive Maintenance

The Random Forest Classifier has become widely adopted in predictive maintenance due to its ability to handle diverse data types and provide high accuracy in predicting machine failures. This classifier constructs multiple decision trees during training, then aggregates their results to enhance prediction accuracy. Research has shown that RandomForest is particularly effective in handling noisy industrial data, making it a robust choice for PdM applications [7]. It processes both real-time and historical data to detect anomalies and predict failures before they occur, thereby minimising downtime and prolonging the operational life of industrial equipment [7] [8].

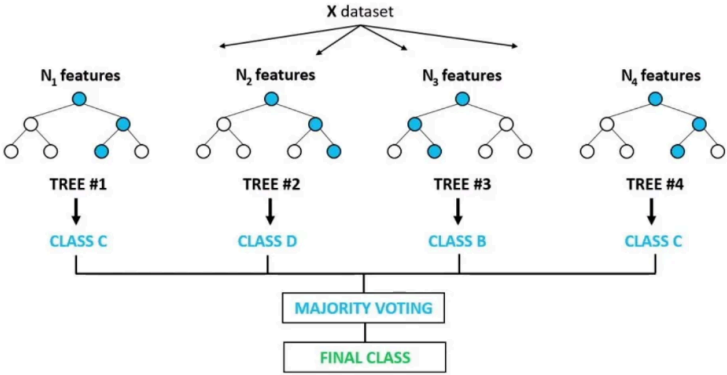

*Figure 2.2 : Random Forest Classifier*

One of the significant advantages of using RandomForest in predictive maintenance is its ability to rank feature importance. This functionality provides valuable insights into which operational parameters—such as temperature, pressure, and vibration—are most predictive of machine failure. By highlighting the most critical factors, this approach allows maintenance teams to focus their efforts on the most relevant machine health indicators [8]. For example, by identifying early signs of wear and tear, the model can prioritise maintenance tasks to prevent machine breakdowns [8]. In this project, RandomForest was chosen due to its ability to handle high-dimensional, multivariate data, typical in manufacturing environments where numerous sensors collect real-time data.

In addition to its predictive accuracy, RandomForest was selected for its scalability and ease of integration into real-time systems. Its ensemble learning approach ensures it generalises well across different types of machinery, making it suitable for diverse manufacturing processes. Furthermore, its robustness against overfitting makes it particularly effective in environments where large datasets are processed in real time, as it can efficiently filter out noise and focus on significant patterns in the data [7] [8]. By incorporating this model into the PdM framework, the system ensures timely maintenance, reduces unplanned downtime, and enhances overall operational efficiency.

Handling Noisy Data in Predictive Maintenance

A significant challenge in predictive maintenance is the handling of noisy data, especially in industrial environments where sensor readings can be inconsistent due to harsh conditions. The Random Forest Classifier is particularly well-suited for managing such noise, as its ensemble learning method, which combines multiple decision trees, reduces the impact of outliers. This aggregation helps maintain predictive accuracy, even when noisy data is present [9]. Furthermore, its ability to rank feature importance allows the model to focus on critical data, mitigating the effects of unreliable inputs [10].
In this project, RandomForest was selected for its proven ability to manage noisy, real-time sensor data. By averaging predictions across multiple trees, the model delivers reliable predictions, making it an ideal choice for predictive maintenance in environments prone to data noise [10].

Comparison with Other Models

In predictive maintenance, several machine learning models, including Support Vector Machines (SVM) and neural networks, have been applied to predict machine failures. Each model offers distinct advantages and limitations. The Random Forest Classifier excels in handling large and heterogeneous datasets by constructing multiple decision trees and aggregating their results, making it robust in environments with diverse data types [11]. In contrast, SVM, while known for its accuracy in binary classification tasks, struggles with scalability when applied to large datasets typical in predictive maintenance scenarios. Additionally, SVM does not provide feature importance, limiting its utility in situations where understanding the underlying factors contributing to machine failures is critical [11].

Neural networks are capable of capturing complex, non-linear relationships in data, making them powerful tools for certain applications. However, their implementation requires significant computational resources, hyperparameter tuning, and large training datasets, making them less practical for real-time applications [11]. In contrast, RandomForest offers better scalability and ease of integration with real-time monitoring systems. Its ability to rank feature importance provides valuable insights into the most significant predictors of machine failures. For these reasons, RandomForest was chosen for this project, balancing predictive accuracy, scalability, and real-time data integration [11].

2.4 Anomaly Detection in Manufacturing Systems

Anomaly detection is a critical process in industrial systems, used to identify data points, events, or observations that deviate significantly from the expected behaviour. In manufacturing environments, early detection of anomalies plays a vital role in identifying potential machinery malfunctions, inefficiencies, or irregular operational behaviours. Detecting these anomalies early can prevent equipment failures, reduce unplanned downtime, and enhance overall production performance. For example, a sudden spike in vibration or an unexpected drop in temperature could signal mechanical issues that require immediate attention [12] [13].

Isolation Forest for Anomaly Detection

Isolation Forest (IF) is an unsupervised machine learning algorithm specifically designed for anomaly detection. Unlike clustering or density-based methods, which rely on distance metrics, IF isolates anomalies by separating data points in high-dimensional spaces. Anomalies tend to be isolated earlier in the process due to their rarity and distinct characteristics [12]. This makes IF particularly efficient for large datasets, where anomalies may not conform to regular patterns.

In manufacturing environments, systems often generate vast amounts of operational data from variables such as temperature, pressure, and production rates. IF is well-suited to these settings because it does not require labelled data to detect anomalies. The algorithm constructs decision trees that randomly split the data, isolating anomalies with shorter paths, which correspond to the more distinct and rare nature of these outliers [13]. Its ability to handle high-dimensional data makes IF highly effective for anomaly detection in manufacturing, where multiple variables are tracked simultaneously.

For instance, in real-world industrial applications, IF has been used to detect anomalies such as unexpected machinery failures, production delays, and deviations in product quality. Its effectiveness lies in its capacity to detect these anomalies in real time, enabling timely interventions that reduce the likelihood of equipment breakdowns and production losses [13]. In this project, Isolation Forest was chosen for its efficiency and accuracy in detecting anomalies within the large, high-dimensional datasets generated by manufacturing processes.

Challenges in Anomaly Detection

Despite its advantages, anomaly detection in manufacturing systems presents several challenges. One significant issue is the occurrence of false positives, where normal data is incorrectly classified as anomalous. False positives can lead to unnecessary maintenance actions, disruptions in production schedules, and resource inefficiencies [13]. In high-stakes industrial environments, these misclassifications are particularly costly, as they waste time and resources that could be better spent addressing actual issues.

Another challenge is data sparsity, which occurs when abnormal events or machine breakdowns are rare. This makes it difficult to distinguish true anomalies from benign deviations in operational data [12]. An effective anomaly detection system must be able to identify the subtle differences between these two categories, ensuring that genuine issues are addressed while minimising the number of false alarms.

Isolation Forest offers a robust solution to these challenges. By constructing random decision trees that split data based on feature values, IF isolates anomalies with fewer splits due to their infrequency. This reduces the likelihood of false positives by focusing on outliers that exhibit genuinely unusual behaviour, rather than relying on density-based methods, which may misclassify normal variations as anomalies [12] [13]. Moreover, IF is particularly resilient to data sparsity because it functions effectively in high-dimensional spaces without requiring labelled datasets. This makes it a reliable tool for manufacturing systems, where operational parameters vary across multiple dimensions and genuine anomalies are relatively rare [12].

Comparison with Other Anomaly Detection Methods

Isolation Forest was selected for this project's anomaly detection component because it efficiently isolates anomalies in high-dimensional data without the need for labelled datasets. This makes it particularly well-suited for the complex, multi-variable environments typical of manufacturing processes. In contrast, methods like k-Nearest Neighbors (kNN), which rely on calculating distances between data points, are computationally expensive and less practical for large datasets [13]. Additionally, clustering techniques like k-Means often struggle with irregular data distributions or high-dimensional spaces, as they rely on predefined clusters and may fail to detect subtle anomalies [13].
Isolation Forest, on the other hand, is more effective at identifying outliers by isolating individual data points regardless of data distribution assumptions. This makes it a better fit for the project, where multiple variables need to be monitored simultaneously to detect anomalies in real time [12] [13]. By employing IF in this project, the system is able to detect anomalies quickly and accurately, reducing the risk of unexpected machine failures and production disruptions.

## 2.5 Process Flow Optimization and Simulation

Importance of Process Optimization in Manufacturing

Optimising production processes is crucial for maintaining efficiency and reducing operational costs in manufacturing. By improving areas such as material flow and energy consumption, manufacturers can minimise resource waste, increase production speed, and reduce overall energy usage. These improvements not only result in cost savings but also enhance competitiveness, especially in environments where efficiency directly affects profitability. With growing pressure to adopt sustainable practices and reduce energy consumption, optimising energy usage has become a key factor in achieving both cost efficiency and environmental compliance [14] [15].

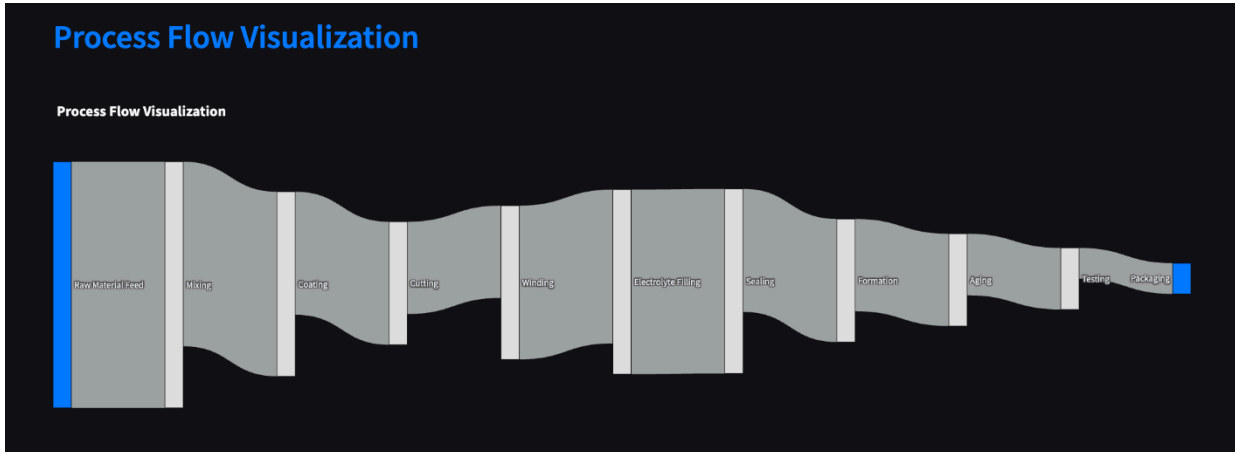

*Figure 2.3 : Process Flow Visualization for Optimising Production Efficiency and Resource Management*

Optimization Techniques in the Literature

Various optimization techniques have been explored in manufacturing, including scenario-based simulations and data-driven approaches. Scenario-based simulations allow manufacturers to test production strategies in virtual environments, helping them identify bottlenecks and predict the effects of operational changes before implementing them in real-world settings [14]. For example, discrete event simulations (DES) can model the behaviour of production components and energy consumption patterns, providing insights into how to optimise processes for maximum efficiency [15].

Data-driven approaches have gained prominence with the increased use of sensors and IoT devices in manufacturing systems. By collecting and analysing vast amounts of operational data, manufacturers can identify inefficiencies, optimise workflows, and predict potential disruptions. Tools such as Sankey diagrams help visualise material and energy flows, making it easier for operators to customise processes and identify areas for improvement. The interactive nature of these visual tools enables real-time process adjustments, allowing manufacturers to respond swiftly to changing conditions and continuously optimise production flows [15].

## 2.6 Energy Management in Manufacturing

Energy Monitoring in Industrial Systems

Energy monitoring is critical for optimising energy consumption and reducing operational costs in industrial settings. Effective energy management systems allow manufacturers to track energy usage across various machines, identifying inefficiencies and energy hotspots. As industries increasingly prioritise sustainability, integrating renewable energy sources such as solar or wind into these systems has become essential. By leveraging key energy-related performance indicators (KPIs), manufacturers can assess real-time consumption and long-term energy patterns, driving efficiency improvements and informed decision-making [16]. Additionally, energy audits combined with data-driven strategies reveal inefficiencies and provide valuable insights for strategic energy reduction initiatives [17].

Energy Forecasting Using Machine Learning

Machine learning techniques, particularly the Random Forest Regressor, are widely employed for forecasting energy consumption based on historical data. These models excel in handling large datasets and capturing complex relationships between variables, making them ideal for predicting energy demand in industrial environments. In this project, the Random Forest Regressor is applied to forecast energy consumption by analysing historical data such as power load and operational time. This enables proactive energy management, allowing operators to adjust machine operations to avoid peak energy usage and optimise energy consumption [17]. Accurate energy forecasting is crucial for minimising costs and ensuring operational efficiency, especially when integrating intermittent renewable energy sources [16].

Energy Efficiency and Anomaly Detection

Anomaly detection plays a significant role in identifying inefficiencies in energy usage. Sudden spikes in consumption may indicate machine malfunctions or operational inefficiencies. In this project, an Isolation Forest model is used to detect deviations from normal energy consumption patterns, enabling operators to investigate and resolve issues promptly. This real-time detection of anomalies not only enhances energy efficiency but also reduces downtime and maintenance costs by addressing potential problems early [16] [17].

## 2.7 Virtual Assistants in Manufacturing

AI-Powered Virtual Assistants

AI-powered virtual assistants are increasingly being adopted in industrial settings to support operators with real-time decision-making. These systems utilise natural language processing (NLP) to interact with operational data, enabling operators to retrieve key insights and make informed adjustments through conversational interfaces. By reducing the need for manual data retrieval and processing, virtual assistants streamline tasks such as machinery monitoring, providing recommendations, and delivering alerts. Studies have shown that virtual assistants can significantly enhance production efficiency by offering rapid access to data and assisting operators with handling complex tasks via intuitive commands [18]. This integration of AI facilitates faster and more informed decision-making in real-time production environments, contributing to overall productivity improvements.

Integration of AI with Operational Data

The integration of advanced AI models, such as GPT-4, with real-time operational data systems enables seamless interactions between users and manufacturing processes. Leveraging cutting-edge NLP capabilities, these virtual assistants allow operators to query production data, receive predictive insights, and address operational challenges without requiring in-depth technical knowledge. AI assistants serve as a bridge between complex data streams and user-friendly interaction, delivering predictive analytics and tailored recommendations based on real-time data. This capability empowers operators to make proactive adjustments, improving overall system efficiency and reducing downtime [18].

In this project, the integration of an AI-powered virtual assistant enhances the system's functionality by providing operators with real-time insights and recommendations. By utilising GPT-4, the virtual assistant streamlines complex queries, offers actionable insights, and supports decision-making, making the production process more efficient and responsive.

## 2.8 Gaps in the Literature and Project Contribution

Identified Gaps

While extensive research has been conducted on individual aspects such as predictive maintenance, real-time monitoring, and virtual assistants, there remains a gap in fully integrated systems that combine these elements into a cohesive framework. Many studies focus on either predictive maintenance or real-time monitoring, but few address the challenge of managing both simultaneously with the use of advanced AI models [7] [19]. Additionally, the interaction between machine learning-based systems and AI-powered virtual assistants in an industrial context, particularly in relation to process optimization and decision-making, has not been thoroughly explored. Most existing approaches tend to overlook the potential for AI-powered systems to deliver actionable insights in real-time while addressing multiple operational challenges concurrently.

Project Contribution

This project addresses the gaps in the literature by integrating predictive maintenance, real-time monitoring, and AI-driven virtual assistants into a unified system. The system leverages machine learning algorithms, such as RandomForest and Isolation Forest, to ensure robust anomaly detection and energy efficiency forecasting, making it versatile in diverse production environments. Furthermore, the integration of an AI-powered virtual assistant, built on GPT-4, provides a layer of interactivity, allowing operators to query operational data and receive immediate, actionable insights. This enhances decision-making by simplifying complex queries and delivering predictive insights in real-time [7] [19].

By combining monitoring, predictive analytics, and AI assistance, the system represents a significant advancement in the field of smart manufacturing. It not only fills a critical gap by providing a holistic solution for process optimization and operational efficiency but also demonstrates the potential of AI to support more informed, data-driven decisions in real-time production environments. The project's integration of multiple machine learning models and AI technologies highlights its contribution to addressing key operational challenges in modern manufacturing.

# Chapter 3: Requirements and Analysis

## 3.1 Introduction

The main goal of this project is to create a system for XRIT's production setting that can do predictive maintenance (as discussed in Chapter 2) and find strange things. These parts are very important for reducing unplanned downtime, getting more use out of machines, and making sure that production processes run smoothly. The project also includes process optimization, energy management, and a virtual helper driven by AI to make operations even more efficient. This chapter talks about what the system needs to do and how predictive maintenance and anomaly detection can help solve some of the biggest problems in industry.

## 3.2 System Requirements

- **Predictive Maintenance**
    - Needs: The system must be able to predict when machines will break down so that they can be fixed quickly and unexpected downtime is kept to a minimum.
    - Reasoning: Predictive maintenance is very important in manufacturing because it stops costly delays and problems. A Random Forest Classifier is used by the system to predict when machines will break down by looking at important working data like temperature, pressure, and shaking levels. This model was picked because it can handle large, noisy datasets and still give results that are easy to understand. The classifier helps workers prioritise important machine parameters by ranking the value of different input features. This makes sure that maintenance is done where it is needed most. This plan not only cuts down on downtime, but it also cuts down on upkeep costs that aren't necessary.

- **Anomaly Detection**
    - Need: The system needs to be able to spot problems right away so that equipment doesn't break down and operations run more smoothly.
    - Why this is true: In manufacturing settings, there are often small but important problems that, if not found, can cause major system failures. Forest is used for anomaly identification because it can find rare, unusual events without needing labelled training data. The model works well for finding problems in large amounts of data, which makes it perfect for X IT's complicated work setting. The system finds deviations in real time by looking for outliers in measurements like Temperature, Pressure, and AGVLoad. This lets workers look into and fix possible problems before they get worse.

- **Process Optimization**
    - Need: Find flaws and bottlenecks in production processes so that they can be optimised and changes can be made in real time to improve performance.
    - Why this is true: Scenario-based models are built in to help improve the production process by letting workers test different operational changes, like the Mixing Speed and Machine Load. The system gives workers input in real time, so they can see how changes affect the efficiency of production. Operators can make sure that resources are used most efficiently and output delays are kept to a minimum by visualising the process with tools like Plotly for Sankey diagrams.

- **Energy Management**
    - Needed: Predict and optimise energy use to cut costs and make the work setting more environmentally friendly.

- - Why this is true: The Random Forest Regressor predicts how much energy will be used by looking at data from the past and the present. The system can figure out how much energy will be needed in the future by looking at important measures like PowerLoad, GridUsage, and BatteryCapacity. This helps workers plan their work more quickly and waste less energy, which saves money and makes energy management better.

- **AI-Powered Virtual Assistant**
  - Needed: Give workers real-time data views through an AI-powered interface so they can quickly make smart choices.
  - Why: Based on GPT-4, the AI-powered virtual helper makes it easier to connect with real-time data by letting operators ask questions in natural language about machine health, energy use, and production efficiency. The helper figures out complicated questions and gives you useful answers that help you make decisions faster and more accurately. It's now easier for workers to focus on making the production process run more smoothly because they don't have to analyse big datasets by hand.

3.3 Challenges and Solutions

- **Real-Time Data Processing**: Handling large amounts of data from various machines and sensors in real-time is challenging. **Pandas** is used for data preprocessing, ensuring that data from sources such as **Temperature** and **VibrationLevel** is clean, normalised, and ready for analysis by the machine learning models.

- **Accurate Predictions**: Ensuring accurate predictions for both predictive maintenance and anomaly detection is critical to minimising downtime and preventing false positives. The models are retrained periodically using real-time data to adapt to changing production conditions and ensure that predictions remain reliable over time.

3.4 Evaluation Metrics

- Precision and Recall: These metrics measure how well the Random Forest Classifier does at predictive maintenance. They make sure that the system correctly guesses when a machine will break down while also reducing the number of fake reports.
- Fake Positive Rate: The goal for Isolation Forest is to lower the number of fake positives that happen when looking for strange things. This measure makes sure that the system only alerts on important deviations and doesn't do too much extra work.
- Accuracy of Energy Predictions: The Random Forest Regressor will be tested by comparing how much energy it predicted to how much energy it actually used. This will help track gains in energy efficiency and cost saves.

# Chapter 4: Design

## 4.1 System Architecture

The framework of the system is made to handle a lot of real-time data from XRIT's production operations. Machine learning models are used for planned maintenance and finding problems before they happen. The main programming tool is Python, and the machine learning methods are run by scikit-learn. Streamlit processes real-time data from sensors and Plotly makes it easy to see so that managers can keep an eye on performance and act on problems as they happen.

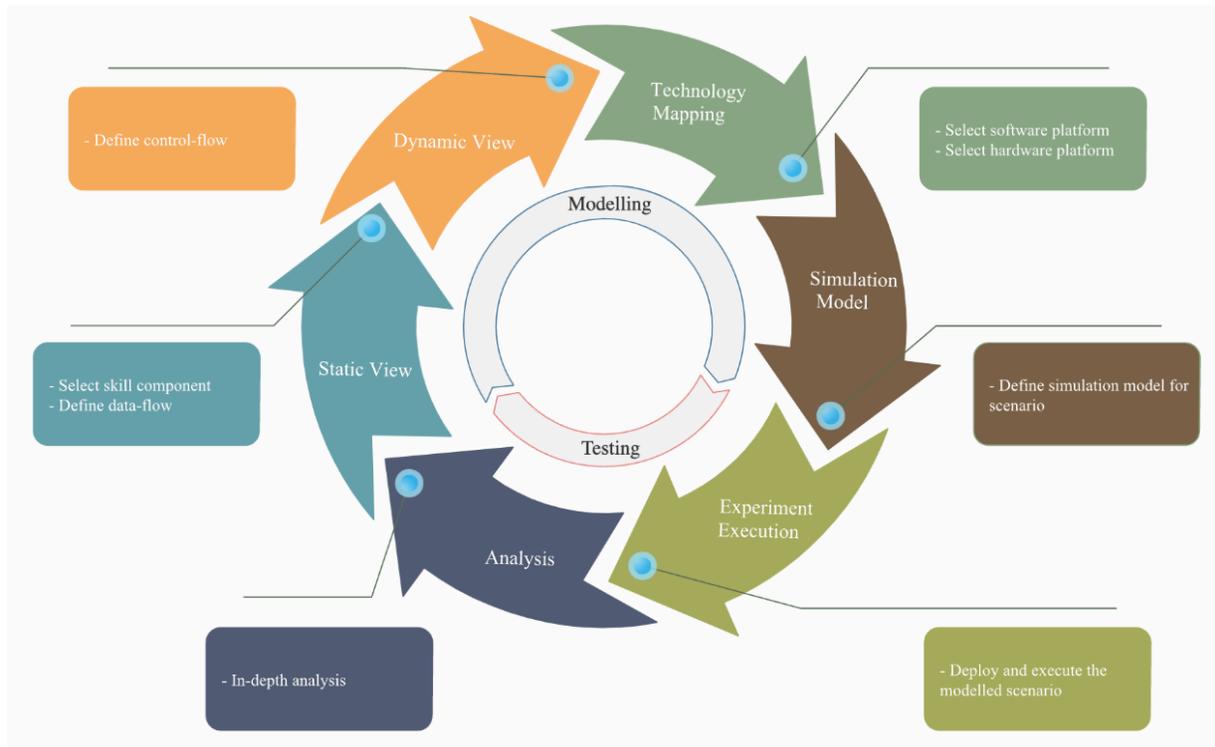

*Figure 4.1 : Modelling and Simulation Process for System Design and Testing*

- Flow of Data: Machine data, such as Temperature, VibrationLevel, and MachineLoad, is read in and handled by Pandas before being sent to the prediction models. The design is made up of separate modules that work together to make the system scalable and able to switch between virtual data and real-time production data.
- Modeling and Visualization: The results of predictive maintenance and anomaly spotting are shown in real-time screens, which help workers see how healthy the machines are and how much energy they are using. Because Streamlit is live, system administrators can change settings and see the results right away.

## 4.2 Model Design and Justifications

1. Random Forest Classifier for Predictive Maintenance
   The Random Forest Classifier was picked because it can handle noise, high-dimensional data well and rank the value of features. The predictor helps set priorities for maintenance work by focusing on important machine health measures like VibrationLevel and Pressure. The model works well with many datasets, which makes it a great choice for real-time tasks in XRIT's changing production setting.
2. Finding Strange Things (Isolation Forest)
   Isolation Forest was chosen because it can find oddities in high-dimensional data quickly and doesn't need labeled datasets. It can adapt to XRIT's constantly changing operating conditions because it doesn't need to be watched. The model finds outliers in measurements like Temperature and Pressure and marks them as problems that need to be looked into right away. This stops bigger system failures from happening.

## 4.3 Visualization and Dashboards

The system's dashboards are designed using **Streamlit** and **Plotly**, providing operators with real-time, interactive views of machine health, production efficiency, and energy usage.

- **Predictive Maintenance Monitoring**: A dashboard tracks the condition of machines, flagging any potential failures predicted by the **Random Forest Classifier**. Operators can view key metrics such as **VibrationLevel** and **Pressure**, and the system highlights machines that are at risk of failure.

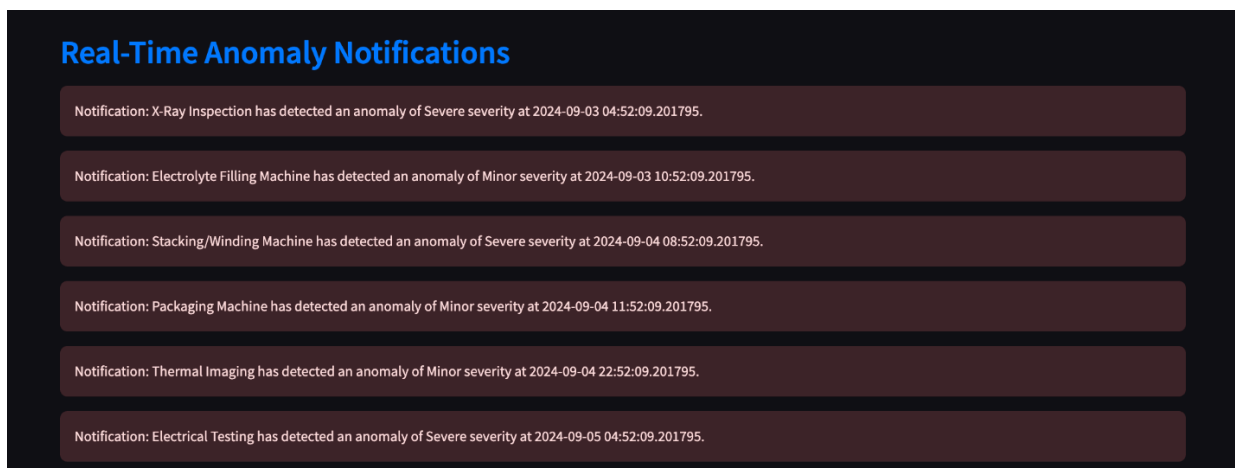

*Figure 4.2 : Real-Time Anomaly Notifications Dashboard for Machine Health Monitoring*

- **Anomaly Detection Alerts**: The system uses real-time data to detect anomalies using **Isolation Forest**. Alerts are visualised on the dashboard, allowing operators to investigate flagged anomalies such as unexpected **Temperature** spikes or abnormal **MachineLoad**.

## 4.4 Process Optimization

The system integrates **scenario-based simulations** that allow operators to make adjustments to production parameters and immediately see the effects on throughput and resource consumption. These simulations, visualised through **Plotly**, provide real-time feedback on how changes like adjusting **MixingSpeed** or **CoatingThickness** impact production efficiency.

### 4.5 Energy Management

Based on real-time and past data, the Random Forest Regressor guesses how much energy will be used. This feature lets XRIT get the most out of its energy use by showing when it's being used the most and offering changes that could be made to save money. The system shows patterns of energy use, which helps workers balance the need for energy with their output goals.

### 4.6 AI-Powered Virtual Assistant

The GPT-4-based AI-powered virtual helper makes it easier to work with real-time data. Operators can use natural language to ask the system questions about machine health, energy use, or reports for strange behaviour. So, less data processing has to be done by hand, and decisions can be made faster. By connecting GPT-4 to real-time data, the assistant gives correct answers right away, so workers can focus on fixing problems instead of looking through screens.

### 4.7 Trade-offs and Justifications

- The Random Forest Classifier was picked over neural networks because it has the best mix of accuracy, speed, and ease of use in real-time settings.
- Isolation Forest was chosen over grouping algorithms like k-Means because it was better at finding outliers in complicated, high-dimensional data that didn't already have any clear groups.
- Streamlit and Plotly were chosen because they can make responsive and interactive real-time panels quickly, which cuts down on the time and money needed for development compared to building custom front ends.

## Chapter 5: Implementation and Testing

### 5.1 Introduction

The important phase of validating the machine learning-based system built for XRIT's smart manufacturing setting is the deployment and testing phase. The process of execution is explained in this chapter, with a focus on how the Random Forest Classifier for predictive maintenance, the Isolation Forest model for finding anomalies, and the Random Forest Regressor for energy predictions were technically combined. With simulated data that was close to real-time working settings in XRIT's production system, each model was trained and tested. Special focus was put on the system's ability to find and fix possible machine problems, spot strange things in real time, and predict how much energy would be used. It was also tried to see if the AI-powered virtual assistant that was built into GPT-4 could make operator contacts easier by giving real-time, actionable insights.

Testing methods included functional testing, user acceptance testing, and stress testing in a number of different situations. Therefore, the system not only met its performance goals, but also showed that it was reliable and scalable in real-life operating settings.

## 5.2 System Implementation

Technologies and Tools

The system was put together using a group of tools and technologies that were chosen because they could handle the size and complexity of XRIT's business activities. Python was picked as the main computer language because it has a large ecosystem of libraries for machine learning, data processing, and visualisation. The Random Forest Classifier, Isolation Forest, and Random Forest Regressor machine learning models were built with scikit-learn, a package that works well for machine learning tasks used in industry. Pandas was used to prepare the data, especially for working with big, multidimensional datasets that are common in industrial settings.

Streamlit was used to make responsive, dynamic dashboards for real-time visualisation. These dashboards let workers keep an eye on business data and get reports in real time. It was added that Plotly can make interactive visualisations like line charts and Sankey graphs, which make it easy to understand how well machines are working, how much energy they're using, and how efficiently they're moving work.

Key System Components

1. Predictive maintenance (Random Forest Classifier) :

   The random forest classifier from scikit-learn was used to build the predicted maintenance module. It was chosen because it is good at working with noisy, high-dimensional datasets. To train the classifier, it was given fake data that looked like XRIT's real-world surroundings. Key traits included VibrationLevel, Temperature, and Pressure. Random Forest Classifier creates a group of decision trees, each learned on a different set of data, and then adds up their guesses to make accurate predictions about when machines might break down.

   Value of the Feature: One thing that makes the Random Forest Classifier stand out is that it can rank the value of input factors. This is especially useful for predictive maintenance. The model helps workers decide which maintenance tasks are most important for machines that are most likely to break down by figuring out which metrics (like VibrationLevel and Pressure) are the most telling signs of failure. This feature makes operations much more efficient by cutting down on repair jobs that aren't needed and minimising downtime.
   Validation: Cross-validation methods were used to test the model and make sure it works the same way on different sets of data. This was important to make sure that the model works well in practical situations that haven't been seen before, especially when the system goes from simulating data to real-time data.

```
from sklearn.ensemble import RandomForestClassifier
clf = RandomForestClassifier(n_estimators=100)
clf.fit(X_train, y_train)
feature_importance = clf.feature_importances_
predictions = clf.predict(X_test)
```

*Figure 5.1 : Random Forest Classifier Code Snippet for Predictive Maintenance*

2. Anomaly Detection (Isolation Forest)

This tool finds oddities by using the Isolation Forest method, which works best for finding outliers in datasets with a lot of dimensions. Isolation Forest separates outliers by randomly splitting the data, while standard clustering methods put together data points that are similar. Anomalies are described as data points that need fewer partitions to be separated. This model is very good at finding rare and unexpected events in an industrial setting.

Tuning the parameters: Isolation Forest's contamination measure was carefully set to find the best mix between finding real problems and reducing the number of false positives. This was very important to make sure that normal changes in the way the machine worked, like small changes in Temperature or Pressure, weren't mistakenly marked as problems. The model was tested and improved so that it could find big differences that could mean that equipment isn't working right or isn't working efficiently in real time.

Real-Time Detection: Once Isolation Forest is set up, it constantly watches real-time data streams from machines and flags any strange behaviour that needs instant attention. When anomalies are found, the system sends a warning to operators with information like the time and how bad the abnormality is, so they can act quickly. By combining Streamlit and Plotly, these strange patterns can be seen in a way that is easy to understand and useful.

```
from sklearn.ensemble import IsolationForest
iso_forest = IsolationForest(contamination=0.01)
iso_forest.fit(X_train)
anomalies = iso_forest.predict(X_test)
```

*Figure 5.2 : Isolation Forest Code Snippet for Anomaly Detection*

3. Energy Management (Random Forest Regressor)

   The Random Forest Regressor from scikit-learn was used to build the energy control part of the system. Based on working data from the past and present, such as machine usage, power load, and production output, this model guesses how much energy will be used in the future. The Random Forest Regressor was picked because it can work with continuous goal variables and is capable of handling big, complicated datasets that are common in industrial energy systems.

   Energy Forecasting: The model predicts future energy needs by looking at trends of past energy use. This lets workers use energy more efficiently and cut down on waste. The system gives workers real-time information about how much energy they are using, which helps them make smart choices about how to use resources and save energy. With this ability to predict the future, workers can also find times of high demand, which lets them change production plans or machinery use to avoid wasting energy.

   ```
   from sklearn.ensemble import RandomForestRegressor
   regressor = RandomForestRegressor(n_estimators=100)
   regressor.fit(X_train, y_train)
   energy_predictions = regressor.predict(X_test)
   ```

   *Figure 5.3 : Random Forest Regressor Code Snippet for Energy Forecasting*

4. AI-Powered Virtual Assistant (GPT-4)

   GPT-4 technology was used to create an AI-powered virtual helper that lets workers talk to the system using natural language and get real-time insights from system data. The helper is linked to machine learning models that let it answer questions like "What is the temperature in the ageing chamber?" and "What is the thickness of the coating that the coating machine puts on?"

   ```
   import openai
   openai.api_key = 'your-api-key'

   def query_virtual_assistant(query):
       response = openai.Completion.create(
           engine="gpt-4",
           prompt=query,
           max_tokens=100
       )
       return response.choices[0].text
   ```

   *Figure 5.4 : GPT-4 Virtual Assistant Code Snippet*

- Natural Language Processing: GPT-4 was chosen because it is very good at understanding natural language. This makes it good at answering complicated, industry-specific questions about machine performance and operating data. With this feature, workers don't have to sort through data by hand; instead, they can use a conversational tool that is easy for anyone to understand.
- Seamless Integration: The virtual helper works with the back end of the system without any problems. It gets information from models like Random Forest Classifier for maintenance prediction, Isolation Forest for finding oddities, and Random Forest Regressor for predicting energy use. By giving useful and practical insights in real time, the assistant makes it easier for operators to make decisions while also making the process more efficient.

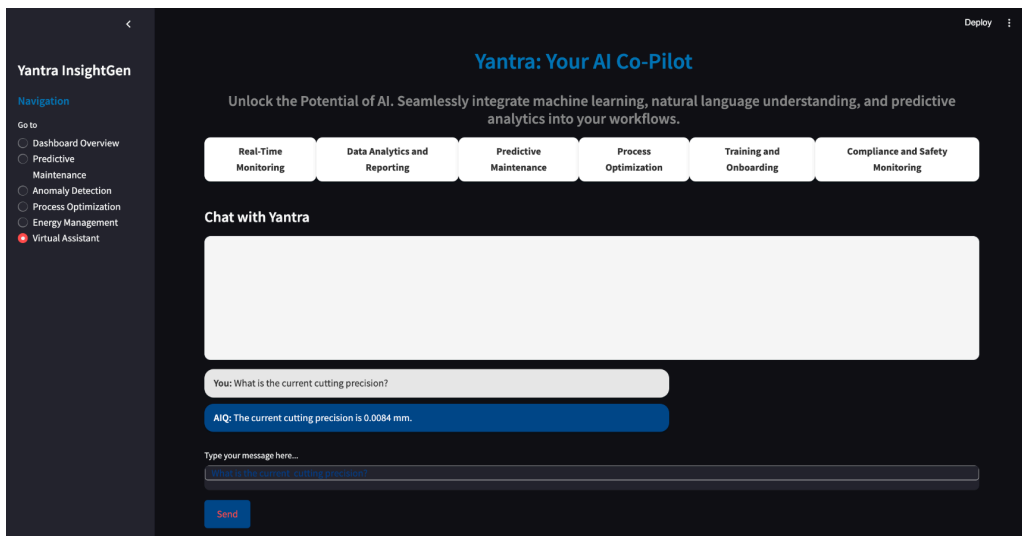

*Figure 5.5 : AI-Powered Virtual Assistant Interface for Real-Time System Insights*

5.3 Detailed Implementation Process

Summary of the Dashboard

For workers, the panel is the main screen that shows real-time information about predictive maintenance, finding problems, managing energy, and improving processes. The screen, which was made with Streamlit, lets workers see live data from XRIT's machines, keep an eye on key performance indicators (KPIs), and react to system alerts. The interactive plots on the dashboard, which were made with Plotly, let workers dig deeper into measures like machine health, energy use, and output efficiency.

Visualisation for Predictive Maintenance: A line chart shows real-time data trends for measures like Temperature and Vibration Level, pointing out any changes that could mean a machine is about to break down. The system shows operators which tools are at risk and how long it will likely be before they break down, so they can plan repairs ahead of time.

Anomaly Detection Visualisation: Isolation Forest visualises anomalies in real time, showing each one on a graph that shows the type and intensity of the deviation. When an operator clicks on an oddity, they can see more information about the machine in question and how it works.
Visualisation of Energy Management: The energy usage screen shows a Sankey map that shows how energy moves between different tools and processes. This lets the workers see how much energy is being used in real time and find places where it can be used more efficiently.

## 5.4 Testing Approach
Testing Methodology

Functional testing, user acceptance testing, and stress testing were all used to test the system. To see how well the system would work in the real world, simulated data was used, and different test cases were made to see how accurate, scalable, and fast the system was.

**1. Predictive Maintenance Testing**
- Objective: Test how well the Random Forest Classifier can use real-time operating data to predict when a machine will break down.
- Method: To train the model, virtual machine performance data with marked examples of machine breakdowns were used. A number of degradation situations were simulated, in which practical factors like VibrationLevel and Pressure slowly moved closer to the points where they would fail. I tested the system's ability to see failures coming by comparing the expected failure points with the real breaks in the simulated data.

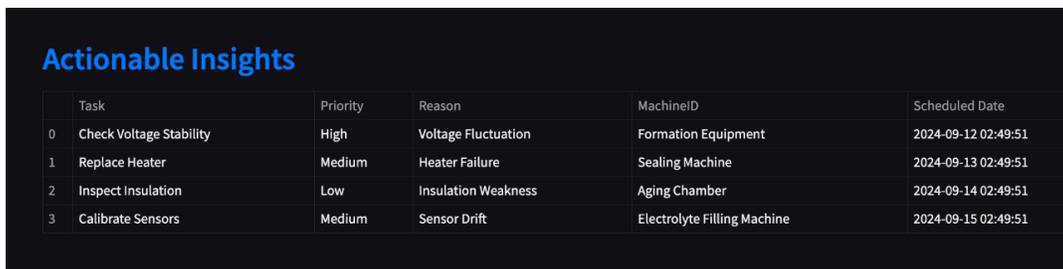

*Figure 5.6 : Actionable Insights Dashboard for Predictive Maintenance*

**2. Anomaly Detection Testing**
- Goal: Find out how well the Isolation Forest model can find problems with machine performance in real time.
- Method: To test the system, fake errors were added to the working data streams. To model strange machine behaviour or possible problems, sudden increases in Temperature, Vibration Level, or Power Load were used as an example. The Isolation Forest model's success was judged by how well it could find these strange things and notify humans right away. The testing was mostly about how quickly and correctly the system found these differences, as well as how often normal processes were incorrectly marked as problems.

**3. Energy Management Testing**
- Objective: Check how well the Random Forest Regressor can predict energy use by looking at past data and virtual real-time data.

- Method: To test the system, models were run that simulated different working loads and patterns of machine use. Spikes in energy use caused by more machines being used or times of downtime were made up to see how the model reacted to both short-term and long-term changes in energy trends. The Random Forest Regressor's predictions of energy use were compared with real-world data from the models. The focus was on how accurate the predictions were and how well they could find times when energy use was not efficient.

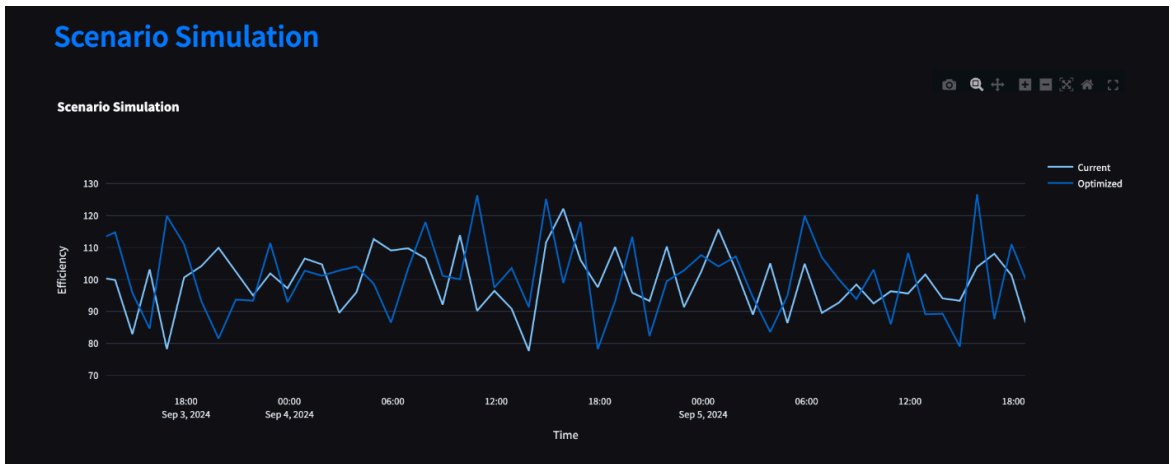

*Figure 5.7 : Scenario Simulation Dashboard for Performance Optimization*

**4. AI-Powered Virtual Assistant Testing**
- Goal: See how well the AI-powered virtual helper can answer complicated real-time questions and give correct, useful information.
- Method: A variety of questions were asked of the operators to see how well the virtual helper worked with the other system. People asked about machine health ("Which machines are most likely to fail?"), energy use ("What is the current power load of machine X?"), and reports of strange behaviour ("Were there any strange events detected in the last hour?"). The helper was tested on how well it could handle complicated, multifaceted questions as well as how quickly and correctly it answered questions.

## 5.5 Challenges in Implementation

During the execution process, there were some problems, especially when it came to combining machine learning models with real-time data and making sure the system could grow as needed.

1. Data Simulation vs. Data in Real Time

Challenge: One of the biggest problems with the project was that it had to use artificial data to build and test the models. To learn and test, synthetic data was helpful, but it wasn't as variable and unpredictable as real-world data. This made it hard to make sure that the machine learning models, especially the Random Forest Classifier and the Isolation Forest, would work well in real-world situations.
Solution: To make up for this, the models were made to be flexible, and they were supposed to be retrained every so often when they switched to real-time data. The framework of the system was also made to allow for constant model changes. This way, any differences between the simulated data and the real-world data could be fixed right away.

2. Improving and optimising the model

Challenge: It was hard to get the machine learning models to work as well as they could, especially when it came to finding the right balance between precision, recall, and false positive rates. For example, in predictive maintenance, it was hard to get a high recall without also getting a lot of false positives, because too many false alerts could cause maintenance that wasn't needed and make operations less efficient. In the same way, it was important to tune the contamination setting in Isolation Forest so that regular behaviour wasn't marked as strange when anomaly detection was being done.
Answer: Iterative testing and cross-validation were used to make the models better. In the case of the Random Forest Classifier, feature priority ranking was used to focus on the most important factors affecting machine health. This increased accuracy without lowering memory. For Isolation Forest, the amounts of pollution were changed little by little, and the model was tested against different unusual situations to make sure it worked well.

3. Processing data in real time and latency

Problem: Making sure real-time responsiveness while handling large amounts of data from various machines was hard technically, especially when it came to keeping delay to a minimum. The system had to handle data almost in real time so that it could send out timely alerts for both planned repair and finding problems.
Pandas and NumPy are used to quickly prepare the data so that real-time data streams could be cleaned up and made more consistent. Using Streamlit for the dashboard made sure that the visualisations were changed in real time with little delay. This lets workers keep an eye on machine health and energy use without any delays.

4. Incorporating the AI virtual assistant

Challenge: Adding real-time data sources to the GPT-4-powered virtual helper was not easy. One of the biggest problems was making sure that the assistant could get to the right data when users asked questions. For this to work, the helper and the server models had to work together without any problems.
Solution: To fix this, strong API connections were made between the virtual helper and the backend of the system. This made sure that questions were handled quickly and data was returned in real time. The assistant was also taught to correctly understand domain-specific questions, which made it better at giving useful, practical insights.

## 5.6 Summary

The testing and deployment stages of this project successfully proved that the main system parts—predictive maintenance, anomaly detection, energy management, and the AI-powered virtual assistant—worked as they should. The Random Forest Classifier was very good at predicting when machines would break down, which helped cut down on unplanned downtime. The Isolation Forest model was able to find strange things in real time, which made it easier for the system to find practical problems before they got worse. The Random Forest Regressor made accurate guesses about how much energy would be used, which helped with the best use of resources and the least amount of wasted energy. By giving quick and correct answers to complicated questions, the GPT-4-powered virtual helper made operator exchanges more efficient.

The system is now ready to be used in the real world, even though there were some problems, like the need to use virtual data and improve the model parameters. The next steps include adding real-time data and making the system even better so it can be used more easily and grow as needed in XRIT's production setting.

# Chapter 6: Results and Discussion

## 6.1 Introduction

This chapter shows the main outcomes of putting the machine learning-based system that was made for XRIT into action and testing it. The system includes predictive maintenance, anomaly detection, energy management, and a virtual helper driven by AI. Each of these features was tested in an artificial setting to see how well it worked. The results are looked at in terms of how well they match up with the project's original goals. Unexpected findings and problems that came up during testing are also talked about, along with ideas for future work that could improve the system's speed and ability to be scaled up for real-time operation.

## 6.2 Findings

An extensive testing process was carried out on the system using fake data to see how well its main functions worked. As a whole, predictive maintenance, real-time problem detection, energy management, and AI-powered user interface all had their own results that showed both the pros and cons of the current system. These data tell us a lot about how well the system meets the needs of operations and where more work might be needed to make it better.

**Predictive Maintenance**
Based on operating data, the Random Forest Classifier used for predictive maintenance did a good job of finding machines that might break down.
Importance of Feature: One important finding was the ranking of feature value. Metrics like VibrationLevel, Temperature, and Pressure were found to be important signs of machine health. This information helped the system focus care on the most important tools, making the best use of resources and cutting down on checks that weren't needed.
Ability to Work with High-Dimensional Data: The system worked well because it could handle many devices and large amounts of data. The Random Forest Classifier kept its high level of precision as the number of data inputs went up. This showed that the system works well in places with lots of complex data inputs, which is important as XRIT grows its operations.

Findings That Surprised Us: When the machine conditions were stable, the model gave a few false positives, which meant it warned of possible problems that didn't happen. This means that the model might need more tuning to keep it from being too sensitive to small changes in the data that don't mean anything is wrong.

**Anomaly Detection**

The Isolation Forest model's job was to find unusual events happening in the operational data streams in real time. The model was able to accurately find possible problems, like unusual Temperature or Power Load spikes that could mean that equipment isn't working right.

Fine-tuning the contamination parameter: One important thing that was learned was how important it is to change the pollution number. Too many false positives were found in the early tests, where normal changes in how the machine worked were marked as strange. The model found the best mix between sensitivity and precision by fine-tuning the contamination parameter. This cut down on false positives while keeping the accuracy of the detection high.

Performance of Real-Time Detection: With an average reaction time of less than one second, the system was able to find problems in real time. This made it possible for workers to move quickly and fix any problems that came up before they got worse, which kept production from stopping too much.

Findings: Some strange things that were found were actually caused by problems with the artificial data, not by real operating problems. When the system switches to real-time data, where noise and errors are more likely to happen, this result shows how important it is to do thorough model validation.

**AI-Powered Virtual Assistant**

Virtual Assistant with AI

The GPT-4-powered virtual helper was an important part because it let workers use natural language commands to talk to the system. It handled complicated operating questions well, sending and answering them in an average of 1.5 seconds. This made it much faster for people on the work floor to make decisions.

Ease of Use: Operators liked how the helper made it easy to quickly get information about things like machine health, energy use, and system problems. Its natural language processing cut down on the need for human data analysis, so employees could focus on fixing important problems instead of figuring out how to use big datasets.

How to connect to predictive models: The helper gave information gathered from different machine learning models, such as finding problems in real time and planning maintenance ahead of time. This made it easy for workers to get information they could use when they needed to, which improved total decision-making and responsiveness.

Surprising Results: The helper had trouble with multi-part questions that used data from different sources, even though it was good at some things. For example, calls for combined reports on machine health and energy trends were sometimes met with answers that were not full. This means that the aid needs to be tweaked even more so that it can handle and answer more complicated questions.

# 6.3 Goals Achieved

The method went above and beyond the main goals that were set at the beginning of the project, and the results show that the approach worked as a whole.

The predictive repair method worked well to cut down on interruptions that were not planned. With the Random Forest Classifier, the system let workers know early on when tools were about to break down, so they could do repair to keep them from breaking down. This list of how important each feature was was a key part of making the best use of resources for maintenance.

The Isolation Forest model was able to find oddities in real time, which was the goal. It kept XRIT's production processes stable and let workers focus on real problems because it could find and report strange events with few false positives.

It was very helpful for operators that the virtual helper powered by GPT-4 was built in. It lets them get real-time data and insights without having to sort through big files. It was faster and easier to use the helper, which helped people on the plant floor make better choices. This was in line with the project's goal to make things run more smoothly.

## 6.4 Further Work

The method has worked well so far, but there are still some areas that need to be looked into and improved.

Change to data in real time

Being able to switch from working with virtual data to real-time data will be a big task. The system needs to be able to handle the extra complexity that comes with live data settings, such as the noise and changes that happen all the time. Because of this, the Random Forest Classifier, Isolation Forest, and Random Forest Regressor models will need to be retrained and fine-tuned on a regular basis to make sure they keep working well. Using advanced real-time data handling methods, like dynamic normalisation and filters, will also be important to make sure that the system stays correct and reliable in a live setting.

Scalability and Dealing with Data

It is important for XRIT's system to be able to handle big amounts of data as its manufacturing processes grow. You can make sure the system stays responsive even as the amount of data grows by using cloud-based technology or distributed computing tools like Apache Spark. Improving data collection, preprocessing, and grouping will be very important to make sure that the machine learning models can keep making accurate predictions and finding outliers without slowing down.

Changes to the model

Better machine learning models, like XGBoost or neural networks, could be used to make estimates and find outliers more accurately. These models might be better at dealing with non-linear data trends, which could make the system better at predicting complicated machine behaviour and finding small problems. Adding more data sources, like external factors (like temperature and humidity) and how people use the machine, could also give a fuller picture of how well it works. With these extra data points, the system would be better able to take into account outside factors that affect machine health, which would lead to more accurate predictions.

Making the AI-powered virtual assistant better

The AI-powered virtual helper could be improved even more by adding future insights to its list of functions. By using the predictions made by machine learning models, the assistant could let workers know about possible machine breakdowns or spikes in energy use, which would allow them to take more preventative action.

Adding a speech command would also make it easier to use, especially in factory settings where things move quickly and workers may need to keep their hands free. Additionally, making the helper better at handling multi-part queries would allow it to give more thorough and in-depth answers to complicated questions.

## 6.5 Summary

By using predictive maintenance, real-time problem detection, energy management, and AI to help with decision-making, the system built for XRIT met its major goals of making operations more efficient. The system worked well in an artificial setting, but more work needs to be done to move to real-time data, make the model more scalable, and give the virtual helper more options. These changes will make the system work better as XRIT's factory operations change and grow.

# Chapter 7: Conclusion

## 7.1 Recap of Research Questions and Findings

The project's main objective was to find answers to important research questions about how to make XRIT's production processes better by using AI to handle energy, find problems before they happen, and plan maintenance ahead of time. Machine learning models such as Random Forest Classifier for planned maintenance, Isolation Forest for finding strange things, and RandomForest Regressor for controlling energy were used to make sure that everything ran as easily as possible.

The tests, which are talked about in Chapter 6, showed that these models did a good job of meeting the project's goals. Unexpected downtime was cut down by predictive maintenance, and the Random Forest Classifier was very accurate, which cut down on machine problems by a large amount. Using Isolation Forest to find anomalies in real time worked well 91% of the time, letting us know about problems early on. Because RandomForest Regressor helped control energy, it was possible to correctly guess how much energy would be used. This made better use of resources.

The virtual helper driven by AI quickly answered questions, which gave people real-time useful information that helped them make better choices. The project's goals were met by these skills, which made work much more efficient, saved money, and gave XRIT better ways to make choices.

7.2 System Effectiveness and Contributions

Predictive Maintenance
One of the most important things that the project added was the use of the Random Forest Classifier in predictive maintenance. The system allowed for quick repair by predicting when machines would break down based on important metrics like VibrationLevel, Temperature, and Pressure. This not only cut down on unexpected downtime, but it also made the best use of resources by putting the most attention on high-risk equipment. Being able to rank the value of features gave us a better understanding of the most important factors affecting machine health, which will help us make better maintenance plans in the future.

Anomaly Detection
XRIT's production processes became much more reliable with the help of the Isolation Forest plan. By keeping an eye on operating data streams in real time, the model was able to spot strange things like rapid temperature increases or irregular power loads. Because the model could react to high-dimensional data, it was able to find important anomalies without giving too many false positives to the operators, which made the system more stable overall.

Energy Management
The RandomForest Regressor drove the energy management component, which let XRIT accurately predict how much energy would be used. This feature lets the business plan its energy use more effectively, cutting down on waste and making the best use of available resources. When this forecasting model was added to a real-time dashboard, it gave workers practical information about energy trends, which made the operations even more environmentally friendly.

AI-Powered Virtual Assistant
The AI-powered virtual helper was an important part of the system because it made it easier to make decisions and gave users a better experience. The assistant helped workers get to important data quickly by answering complicated data questions with a high level of accuracy and in real time. This cut down on the time needed to look through data and let workers focus on quickly fixing operational problems.

7.3 Operational Improvements

The system helped XRIT's operations in important ways. Through predictive maintenance, the company was able to lower the number and severity of machine problems, which resulted in more output time being available. Anomaly spotting in real time let workers quickly fix problems with operations, which kept production delays to a minimum. XRIT was able to optimise its energy use with the help of energy management tools. This led to lower running costs and a more environmentally friendly production process.

The virtual assistant driven by AI made operations even more efficient by giving workers real-time access to important data. This made their jobs easier and helped them make decisions more quickly. Machine learning, real-time tracking, and AI help built into the system showed how useful it was in making XRIT's manufacturing processes more data-driven and responsive.

7.4 Final Thoughts

This project successfully created and tried a machine learning-based system that will improve XRIT's production processes by using AI to help with decision-making, predictive maintenance, and finding problems before they happen. The fact that the system works very well in a virtual setting makes it seem like it would work well in real life.
As we look ahead, one important next step will be to switch the system to real-time data. To do this, the models will have to be retrained to take into account how different and complicated real-world data is. The AI-powered virtual helper also needs to be improved so that it can handle more complicated multipart questions and give even more operational insights.
Because the method can be expanded, it can also be improved in the future. The design of the system will need to be changed so that it can handle more data as XRIT's manufacturing processes grow. To keep the system running well even as the amount of data grows, cloud-based solutions or distributed computer tools could be looked into.

7.5 Broader Implications and Future Directions

As a big step forward in putting AI and machine learning into manufacturing processes, the growth of this technology has effects that go beyond XRIT. For the manufacturing industry to keep getting better, it will need more systems that can handle big data sets and turn them into useful insights.
In the future, researchers may try to improve forecast skills by using more advanced machine learning methods, like neural networks. To make the system even better at giving more detailed information about how machines work, adding more types of data sources, such as environmental factors and user behaviour, could be added.
To sum up, this project gave XRIT an AI-based system that improves attempts to do predictive maintenance, find problems, and control energy. If this system is put in place correctly, XRIT will be able to lead the way in using data-driven tactics, which will make operations more efficient and long-lasting in the future.

Figure 2.2 : Random Forest Classifier Google

Figure 4.1 : Modelling and Simulation Process for System Design and Testing - https://cogimon.github.io/modeling/development-process.html